\documentclass[letterpaper, 10 pt, conference]{ieeeconf}
\IEEEoverridecommandlockouts
\PassOptionsToPackage{hyphens}{url}\usepackage[bookmarks=true]{hyperref}
\usepackage{amssymb}
\usepackage{amsmath}
\usepackage{graphicx}
\usepackage{url}
\usepackage{array}
\usepackage{multirow}
\usepackage{xcolor,colortbl}
\usepackage[ruled,linesnumbered, noend]{algorithm2e}

\definecolor{Gray}{gray}{0.65}
\newcolumntype{a}{>{\columncolor{Gray}}c}

\newenvironment{myitem}{\begin{list}{$\bullet$}
{\setlength{\itemsep}{-0pt}
\setlength{\topsep}{0pt}
\setlength{\labelwidth}{0pt}
\setlength{\leftmargin}{10pt}
\setlength{\parsep}{-0pt}
\setlength{\itemsep}{0pt}
\setlength{\partopsep}{0pt}}}%
{\end{list}}

\title{\LARGE \bf A Self-supervised Learning System for Object Detection using\\ Physics Simulation and Multi-view Pose Estimation}
\author{Chaitanya Mitash, Kostas E. Bekris and Abdeslam Boularias
\thanks{The authors are with the Computer Science Department of
  Rutgers University in Piscataway, New Jersey, 08854, USA. Email:
  \{cm1074,kb572,ab1544\}@rutgers.edu.}
\thanks{This work is supported by NSF awards IIS-1734492, IIS-1723869
  and IIS-1451737. Any opinions or findings expressed in this paper do
  not necessarily reflect the views of the sponsors. The authors would
  like to thank the anonymous IROS'17 reviewers for their constructive
  comments.}%
}

\begin{document}
\maketitle
\thispagestyle{empty}
\pagestyle{empty}

\begin{abstract}
Progress has been achieved recently in object detection given
advancements in deep learning. Nevertheless, such tools typically
require a large amount of training data and significant manual effort
to label objects. This limits their applicability in robotics, where
solutions must scale to a large number of objects and variety of
conditions. This work proposes an autonomous process for training a
Convolutional Neural Network ({\tt CNN}) for object detection and pose
estimation in robotic setups. The focus is on detecting objects placed
in cluttered, tight environments, such as a shelf with multiple
objects. In particular, given access to 3D object models, several
aspects of the environment are physically simulated.  The models are
placed in physically realistic poses with respect to their environment
to generate a labeled synthetic dataset. To further improve object
detection, the network self-trains over real images that are labeled
using a robust multi-view pose estimation process.  The proposed
training process is evaluated on several existing datasets and on a
dataset collected for this paper with a Motoman robotic arm. Results
show that the proposed approach outperforms popular training processes
relying on synthetic - but not physically realistic - data and manual
annotation. The key contributions are the incorporation of physical
reasoning in the synthetic data generation process and the automation
of the annotation process over real images.

\end{abstract}

\section{Introduction}
\label{sec:intro}

\begin{figure}
\centering \includegraphics[width=\linewidth, height=10cm,
  keepaspectratio]{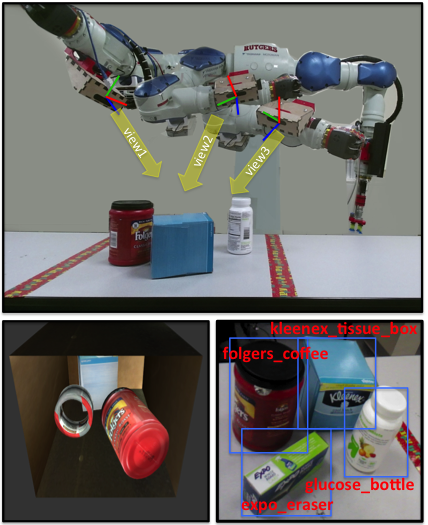}
\vspace{-.1in}
\caption{(top) A robotic arm performs pose estimation from
  multiple view points using an object detector trained in
  physically-simulated scenes (bottom-left). The estimated poses are
  used to automatically label real images (bottom-right). They are
  added to the training dataset as part of a lifelong learning
  process.  Initially, the multi-view pose estimation procedure
  bootstraps its accuracy by trusting the objects' labels as predicted
  from the detector given training over synthetic images. It then uses
  these labels to annotate images of the same scene taken from more
  complex viewpoints.}
\vspace{-.25in}
\label{fig:intro}
\end{figure}

Object detection and pose estimation is frequently the first step of
robotic manipulation. Recently, deep learning methods, such as those
employing Convolutional Neural Networks ({\tt CNN}s), have become the
standard tool for object detection, outperforming alternatives in
object recognition benchmarks. These desirable results are typically
obtained by training {\tt CNN}s using datasets that involve a very
large number of labeled images, as in the case of ImageNet
\cite{ILSVRC15}). Creating such large datasets requires intensive
human labor. Furthermore, as these datasets are general-purpose, one
needs to create new datasets for specific object categories and
environmental setups that may be of importance to robotics, such as
warehouse management and logistics.

The recent Amazon Picking Challenge ({\tt APC}) \cite{Correll:2016aa}
has reinforced this realization and has led into the development of
datasets specifically for the detection of objects inside shelving
units \cite{singh2014bigbird,rennie2016dataset,Princeton}. These
datasets are created either with human annotation or by incrementally
placing one object in the scene and using foreground masking. \par

An increasingly popular approach to avoid manual labeling is to use
synthetic datasets generated by rendering 3D CAD models of objects
with different viewpoints. Synthetic datasets have been used to train
{\tt CNNs} for object detection \cite{peng2015learning} and viewpoint
estimation \cite{su2015render}. One major challenge in using synthetic
data is the inherent difference between virtual training examples and
real testing data. For this reason, there is considerable interest in
studying the impact of texture, lighting, and shape to address this
disparity \cite{SunVirtual}. Another issue with synthetic images
generated from rendering engines is that they display objects in poses
that are not necessarily physically realistic. Moreover, occlusions
are usually treated in a rather naive manner, i.e., by applying
cropping, or pasting rectangular patches, which again results in
unrealistic scenes
\cite{peng2015learning,su2015render,movshovitz2016useful}.

This work proposes an automated system for generating and labeling
datasets for training {\tt CNN}s. The objective of the proposed system
is to reduce manual effort in generating data and to increase the
accuracy of bounding-box-based object detection for robotic setups. In
particular, the two main contributions of this work are:

\begin{itemize}
\item A physics-based simulation tool, which uses information from
  camera calibration, object models, shelf or table localization to
  setup an environment for generating training data. The tool performs
  physics simulation to place objects at realistic configurations and
  renders images of scenes to generate a synthetic dataset to train an
  object detector.
\item A lifelong, self-learning process, which employs the object
  detector trained with the above physics-based simulation tool to
  perform a robust multi-view pose estimation with a robotic
  manipulator, and use the results to correctly label real images in
  all the different views. The key insight behind this system is the
  fact that the robot can often find a good viewing angle that allows
  the detector to accurately label the object and estimate its
  pose. The object's predicted pose is then used to label images of
  the same scene taken from more difficult views, as shown in
  Fig.~\ref{fig:intro}. The transformations between different views
  are known because they are obtained by moving the robotic
  manipulator.
\end{itemize}
The software and data of the proposed system, in addition to all the
experiments, are publicly available at
\url{http://www.physimpose.com}

\begin{figure*}
\centering
\includegraphics[width=\textwidth, height=7.5cm,
  keepaspectratio]{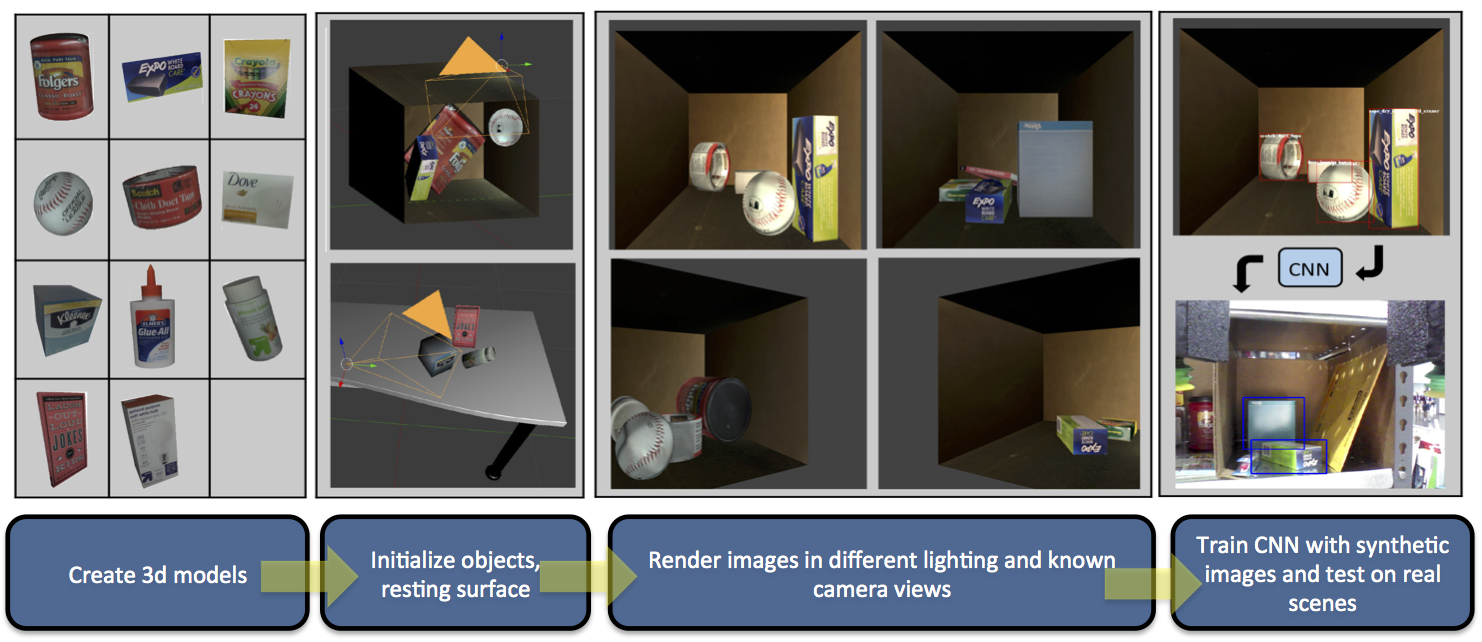}
\vspace{-.15in}
\caption{Pipeline for physics aware simulation: The 3D CAD models are
  generated and loaded in a calibrated environment on the simulator. A
  subset of the objects is chosen for generating a scene. Objects are
  physically simulated until they settle on the resting surface under
  the effect of gravity. The scenes are rendered from known camera
  poses. Perspective projection is used to compute 2D bounding boxes
  for each object. The labeled scenes are used to train a Faster-RCNN
  object detector \cite{ren2015faster}, which is tested on a real-world
  setup.}
\label{fig:physim}
\vspace{-.2in}
\end{figure*}

\section{Related Work}
\label{sec:related}
The novelty of the proposed system lies on the training process for
generating synthetic data as well as augmenting the synthetic data
with real ones that are generated from an automated, self-learning
process. This involves several modules, which have been studied in the
related literature over the years.

\textbf{Object Segmentation:} The tasks of object
detection and semantic segmentation of images have been studied
extensively and evaluated on large scale image datasets. Recently, the
{\tt RCNN} approach combined region proposals with convolutional
neural networks \cite{girshick2014rich}. This opened the path to high
accuracy object detection, which was followed up by deep network
architectures \cite{he2016deep, simonyan2014very} and end-to-end
training frameworks \cite{girshick2015fast,ren2015faster}. There has
also been a significant success in semantic labeling of images with
the advent of Fully Convolutional networks ({\tt
FCN}) \cite{shelhamer2016fully} and its
extensions \cite{chen2014semantic, dai2015boxsup, li2016fully}. This
work utilizes {\tt FCN} and {\tt Faster-RCNN} and proposes an
automated way to collect data and incrementally train the structures
for improved performance.

\textbf{Pose Estimation:} One way to approach this challenge is through
matching local features, such as SIFT \cite{pauwels2015simtrack}, or
by extracting templates using color gradient and surface normals from
3D object models \cite{hinterstoisser2012model}. Synthesis-based
approaches have also been gaining
popularity \cite{krull2015learning,d2p}. Nevertheless, in application
domains, such as those studied by the Amazon Picking
Challenge \cite{Correll:2016aa}, which involve varying light
conditions and cluttered scenes, it has been shown \cite{Princeton}
that {\tt CNN}-based
segmentation \cite{ren2015faster,shelhamer2016fully} followed by point
cloud registration with 3D models \cite{icp,super4pcs,koltun} is an
effective approach. This paper builds on top of these techniques for
pose estimation and proposes a method to self-feed the output of such
processes to improve accuracy.

\textbf{Synthetic Datasets:} Synthetic datasets generated from 3D
models have been used for object detection \cite{peng2015learning,
stark2010back} and pose estimation \cite{gupta2015aligning,
su2015render} with mixed success as indicated by an evaluation of the
performance of detectors trained on synthetic images to those trained
with natural images \cite{movshovitz2016useful}. This work proposes
the incorporation of a physics-based simulator to generate realistic
images of scenes, which helps object detection success rate.

\textbf{Self-supervised Learning:} The idea of incrementally learning
with minimal supervision has been exploited previously in many
different ways. Curriculum learning \cite{bengio2009curriculum} and
self-paced learning \cite{kumar2010self} have been adapted to improve
the performance of object detectors \cite{liang2015towards,
sangineto2016self}. The self-learning technique proposed here involves
the robot acquiring real images of scenes from multiple views. Then
the robot uses the knowledge acquired from confidently detected views
and 3D model registration to improve object detection in a life-long
manner.

\section{Physics-aware Synthetic Data Generation}
\label{sec:physim}
The proposed system starts by physically simulating a scene as well as
simulating the parameters of a known camera. The accompanying tool
generates a synthetic dataset for training an object detector, given
3D CAD models of objects. This module has been implemented using the
Blender API \cite{Blender}, which internally uses the Bullet physics
engine \cite{Bullet}. The pipeline for this process is depicted in
Fig. \ref{fig:physim}, while the corresponding pseudocode is provided
in Alg.~\ref{alg:alg1}. The method receives as input:

\begin{myitem}

\item a set of predefined camera poses ${\bf P}_{cam}$,

\item the pose of the resting surface ${\bf P}_{s}$,

\item the intrinsic parameters of the camera $\textbf{C}_{int}$,

\item the set of 3D object models {\bf M} and

\item the number of simulated training images $N$ to generate.

\end{myitem}

\vspace{-.15in}

\begin{algorithm}[h]
\caption{{\sc physim\_cnn}$( {\bf P}_{cam}, {\bf P}_{s}, \textbf{C}_{int}, {\bf M}, N)$ }
\label{alg:alg1}
\textbf{dataset} $\gets \emptyset$\;
\While{ $( |{\bf dataset}| < N )$ }
{
    \textbf{O} $\gets$ a random subset of
    objects from {\bf M}\;
    \textbf{P}$^{O}_{init} \gets $ {\sc initial\_random\_poses}( \textbf{O} )\;
    \textbf{P}$^{O}_{final} \gets$ {\sc phys\_sim}(
    {\bf P}$^{O}_{init}$, {\bf P}$_{s}$, {\bf O})\;
    {\bf Light} $\gets$ {\sc pick\_lighting\_conditions}()\;
    \ForEach { $({\bf view} \in {\bf P}_{cam})$ }
    {
        \textbf{image} $\gets$ {\sc render}( {\bf
    P}$^{O}_{final}$, {\bf view}, C$_{int}$, {\bf Light})\;
	\{ \textbf{labels, bboxs} \} $\gets$ {\sc Project}({\bf P}$^O_{final}$,
    {\bf view})\;
        {\bf dataset} $\gets$ {\bf dataset} $\cup$ ({\bf image, labels, bboxs});
    }
}
{\sc sim\_detect($\cdot$)} $\gets$ {\sc faster-rcnn}( {\bf
    dataset} )\;
\Return{\sc sim\_detect($\cdot$) and {\bf dataset}}\;
\end{algorithm}

\vspace{-.15in}

In a sensing system for robotic manipulation, a 6 degree-of-freedom
(DoF) pose of the camera mounted on a robotic arm $({\bf view} \in
{\bf P}_{cam})$, can be exactly computed using forward
kinematics. Furthermore, the camera calibration provides the intrinsic
parameters of the camera $\textbf{C}_{int}$.  To position the resting
surface for the objects, a localization process is performed first in
the real-world to compute the pose of the resting surface ${\bf
P}_s$. The system has been evaluated on an {\tt APC} shelf and a
table-top environment. The shelf localization process uses {\tt
RANSAC} \cite{RANSAC} to compute edges and planes on the shelf and the
geometry of the shelf model is used to localize the bins.  Given the
above information as well as 3D object models {\bf M}, the method aims
to render and automatically label $N$ different images in simulation.

The algorithm simulates a scene by first choosing the objects ${\bf
O}$ from the list of available object models ${\bf M}$ (line 3). The
initial pose of an object is provided by function {\sc
initial\_random\_poses} (line 4), which samples uniformly at random
along the x and y axis from the range
$(\frac{-dim_i}{2}, \frac{dim_i}{2})$, where $dim_i$ is the dimension
of the resting surface along the $i^{th}$ axis. The initial position
along the z-axis is fixed and can be adjusted to either simulate
dropping or placing. The initial orientation is sampled appropriately
in {\tt SO(3)}. Then the function {\sc phys\_sim} is called (line 5),
which physically simualates the objects and allows them to fall due to
gravity, bounce, and collide with each other as well as with the
resting surface. Any inter-penetrations among objects or with the
surface are treated by the physics engine. The final poses of the
objects ${\bf P}_{final}^{O}$, when they stabilize, resemble
real-world poses. Gravity, friction coefficients and mass parameters
are set at similar values globally and damping parameters are set to
the maximum to promote fast stabilization.

The environment lighting and point light sources are varied with
respect to location, intensity and color for each rendering (line
6). Simulating different indoor lighting sources according to related
work ~\cite{ReproducingRealWorldLight} helps to avoid over-fitting to
a specific texture. This makes the training set more robust to
different testing scenarios. Once lighting conditions ${\bf Light}$
are chosen, the simulated scene is rendered from multiple views using
the pre-defined camera poses (line 6). The rendering function {\sc
render} requires the set of stabilized object poses ${\bf
P}_{final}^{O}$, the camera viewpoint ${\bf view}$ as well as the
selected lighting conditions ${\bf Light}$ and intrinsic camera
parameters $\textbf{C}_{int}$ (line 7).  Finally, perspective
projection is applied to obtain 2D bounding box labels for each object
in the scene with function {\sc project} (line 8). The overlapping
portion of the bounding boxes for the object that is further away from
the camera is pruned.

The synthetic {\bf dataset} generated is used to train an object
detector {\sc sim\_detect}($\cdot$) based on {\tt
Faster-RCNN} \cite{ren2015faster}, which utilizes a deep {\tt VGG}
network architecture \cite{simonyan2014very}.

\begin{figure*}[thpb]
\centering
\includegraphics[width=\textwidth, height=8cm]{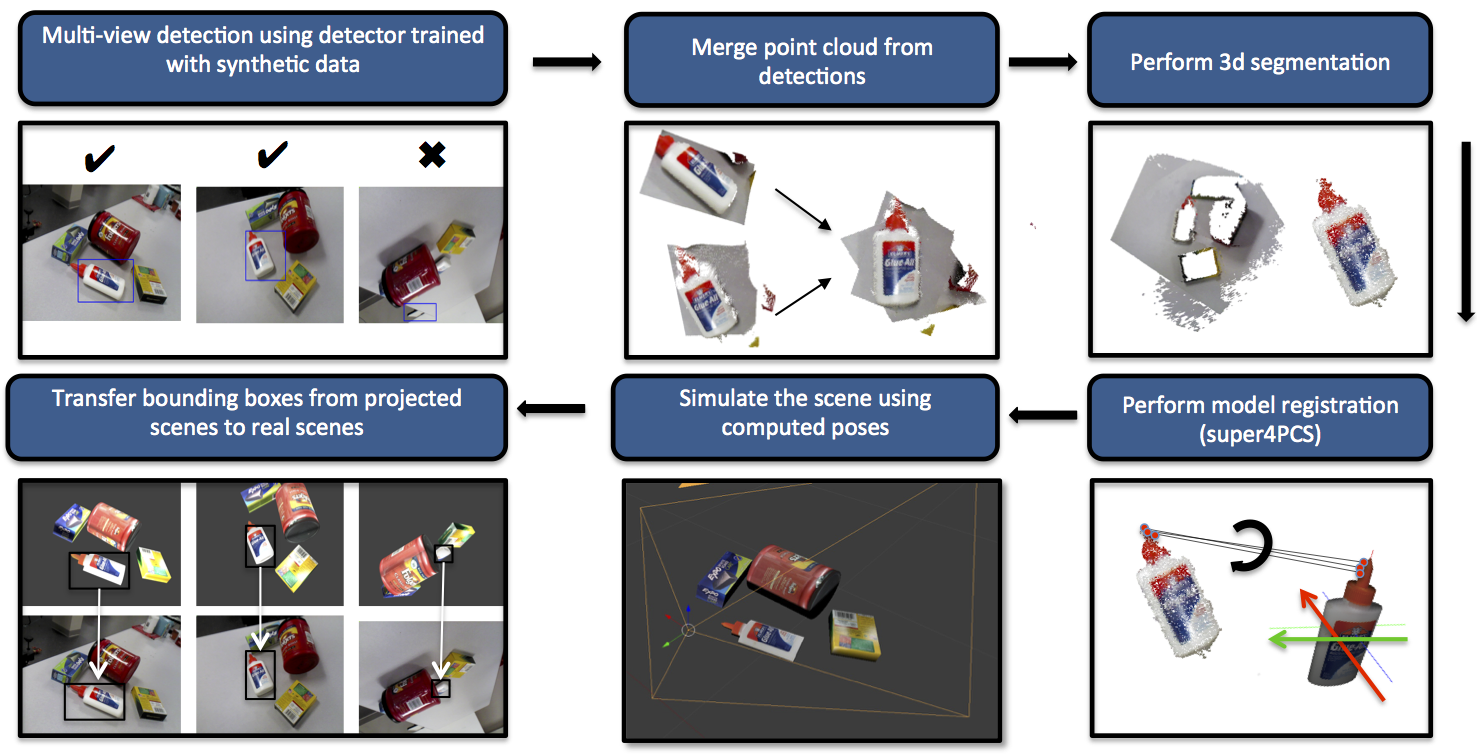}
\vspace{-.3in}
\caption{Automatic self-labeling  pipeline:
The detector, which is trained with simulated data, is used to detect
objects from multiple views. The point cloud aggregated from
successful detections undergoes 3D segmentation. Then,
Super4PCS \cite{super4pcs} is used to estimate the 6D pose of the
object in the world frame. The computed poses with high confidence are
simulated and projected back to the multiple views to obtain precise
labels over real images.}
\label{fig:selflearn}
\vspace{-.2in}
\end{figure*}

\section{Self-Learning via\\ Multi-view Pose Estimation}
\label{sec:self_learning}
Given access to an object detector trained with the physics-based
simulator, the self-learning pipeline labels real-world images using a
robust multi-view pose estimation. This is based on the idea that the
detector performs well on some views, while might be imprecise or fail
in other views. Aggregating 3D data over the confident detections and
with access to the knowledge of the environment, a 3D segment can be
extracted for each object instance in the scene. This process combined
with the fact that 3D models of objects are available, makes it highly
likely to estimate correct 6DoF pose of objects given enough views and
search time. The results of pose estimation are then projected back to
the multiple views, and used to label real images. These examples are
very effective to reduce the confusion in the classifier for novel
views. The process also autonomously reconfigures the scene using
manipulation actions to apply the labeling process iteratively over
time on different scenes, thus generating a labeled dataset which is
used to re-train the object detector. The pipeline of the process is
presented in Fig.\ref{fig:selflearn} and the pseudocode is provided in
Alg.~\ref{alg:alg2}.

A robotic arm is used to move the sensor to different pre-defined
camera configurations ${\bf P}_{cam}$ and capture color ({\bf RGB})
and depth ({\bf D}) images of the scene (lines 2-3). The PRACSYS
motion planning library \cite{kimmel2012pracsys, Littlefield:2015aa}
was used to control the robot in the accompanying implementation.

\vspace{-.15in}
\begin{algorithm}[h]
\caption{{\sc self-learn}$( {\bf dataset}, {\bf P}_{cam}, {\bf M}, N^{\prime}$) }
\label{alg:alg2}
\While{$|dataset| < N^{\prime} $}
{
   \ForEach{ ${\bf view} \in {\bf P}_{cam}$ }
   {
     \{\textbf{RGB$_{view}$, D$_{view}$}\} $\gets$ {\sc capture}( {\bf view})\;
   }      
  \ForEach{ object ${\bf o}$ in the scene and ${\bf M}$ }
  {
    \textbf{Cloud[o]} $= \emptyset$\;
    \ForEach{ {\bf view} $\in$ {\bf P}$_{cam}$}
    {
      {\bf bbox} $\gets$ {\sc sim\_detect}( RGB$_{view}$ )\;
      \If{ {\sc confidence}({\bf bbox}) $>$ threshold }
      {
        \textbf{3DPts} $\gets$ {\sc get\_3DPts}( {\bf bbox},
        {\bf D$_{view}$} )\;
        \textbf{Cloud[o]} $\gets$ \textbf{Cloud[o]} $\cup$ \textbf{3DPts}\;
      }
    }
    {\sc Outlier\_Removal}( {\bf Cloud[o]} )\;
    {\bf P[o]} $\gets$ {\sc Super4PCS}( {\bf Cloud[o]}, {\bf M[o]} )\;
  }
  \ForEach{ ${\bf view} \in {\bf P}_{cam}$}
  {
    \{ \textbf{labels, bboxs} \} $\gets$  {\sc project}$( \bf P[o], view
    )$\;
    {\bf dataset} $\gets$ {\bf dataset} $\cup$ ({\bf RGB}$_{view}$,
    {\bf labels}, {\bf bboxs})\;
  }
  \textbf{randObj} $\gets$ {\sc sample\_random\_object}( {\bf M} )\;
  \textbf{objConfig} $\gets$ {\sc pick\_config}()\;
  {\sc reconfigure\_object}( {\bf randObj}, {\bf objConfig} )\;
}
{\sc new\_detect($\cdot$)} $\gets$ {\sc faster-rcnn}( {\bf dataset} )\;
\Return{\sc new\_detect($\cdot$)}\;
\end{algorithm}

\vspace{-.15in}

The detector trained using physics-aware simulation is then used to
extract bounding boxes {\bf bbox} corresponding to each object $o$ in
the scene (line 7). There might exist a bias in simulation either with
respect to texture or poses, which can lead to imprecise bounding
boxes or complete failure in certain views. For the detection to be
considered for further processing, a threshold is considered on the
confidence value returned by {\tt RCNN} (line 8).

The pixel-wise depth information {\bf 3DPts} within the confidently
detected bounding boxes {\bf bbox} (line 9) is aggregated in a common
point cloud per object {\bf Cloud[o]} given information from multiple
views (line 10). The process employs environmental knowledge to clean
the aggregated point cloud (line 11).  points outside the resting
surface bounds are removed and outlier removal is performed based on
k-nearest neighbors and a uniform grid filter.

Several point cloud registration methods were tested for registering
the 3D model {\bf M[o]} with the corresponding segmented point cloud
{\bf Cloud[o]} (line 12). This included {\sc
Super4PCS} \cite{super4pcs}, fast global registration \cite{koltun}
and simply using the principal component analysis ({\tt PCA}) with
Iterative Closest Point ({\tt ICP}) \cite{icp}. The {\sc Super4PCS}
algorithm \cite{super4pcs} used alongside ICP was found to be the most
applicable for the target setup as it is the most robust to outliers
and returns a very natural metric for confidence evaluation. {\sc
Super4PCS} returns the best rigid alignment according to the Largest
Common Pointset ({\tt LCP}). The algorithm searches for the best
score, using transformations obtained from four-point
congruences. Thus, given enough time, it generates the optimal
alignment with respect to the extracted segment.

After the 6DoF pose is computed for each object, the scene is
recreated in the simulator using object models placed at the pose {\bf
P[o]} and projected to the known camera views (line 14). Bounding
boxes are computed on the simulated setup and transferred to the real
images. This gives precise bounding box labels for real images in all
the views (line 15).

\begin{figure}[t]
\centering
\includegraphics[width=\linewidth]{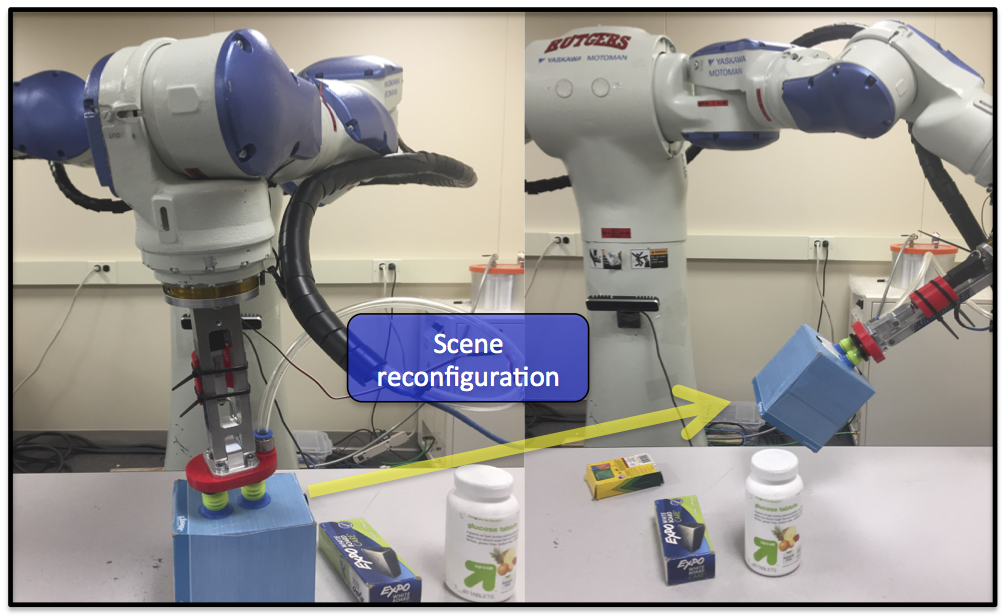}
\vspace{-.3in}
\caption{Manipulator performing scene reconfiguration by moving an object from one configuration on the table to another}
\label{fig:figurelabelx}
\vspace{-.1in}
\end{figure}

To further reduce manual labeling effort, an autonomous scene
reconfiguration is performed (lines 16-18). The robot reconfigures the
scene with a pick and place manipulation action to iteratively
construct scenes and label them, as in
Fig.~\ref{fig:figurelabelx}. For each reconfiguration, the object to
be moved is chosen randomly and the final configuration is selected
from a set of pre-defined configurations in the workspace.

\section{Evaluation}
\label{sec:evaluation}
This section discusses the datasets considered, it compares different
techniques for generating synthetic data and evaluates the effect of
self-learning. It finally applies the trained detector on the 6DoF
pose estimation task. The standard Intersection-Over-Union ({\tt IoU})
metric is employed to evaluate performance in the object detection
task.

\subsection{Datasets}
Several RGB-D datasets have been released in the setting of the Amazon
Picking Challenge \cite{singh2014bigbird, rennie2016dataset,
  Princeton}. They proposed system was evaluated on the benchmark
dataset released by Team MIT-Princeton called the Shelf\&Tote dataset
\cite{Princeton}. The experiments are performed on 148 scenes in the
shelf environment with different lighting and clutter conditions. The
scenes include 11 objects used in APC with 2220 images and 229 unique
object poses. The objects were chosen to represent different geometric
shapes but ignoring the ones which did not have enough depth
information. Thus, the results can be generalized to a large set of
objects.

The proposed system has been also evaluated on a real-world table-top
setup. The corresponding test dataset was generated by placing
multiple objects in different configurations on a table-top. An {\tt
  Intel RealSense} camera mounted on a {\tt Motoman} robot was used to
capture images of scenes from multiple views. Images corresponding to
41 cluttered scenes, with 11 {\tt APC} objects and 473 detection
instances were collected and manually labeled.

\subsection{Evaluating the Object Detector trained in Simulation}
To study how object pose distribution effects the training process,
different techniques for synthetic data generation are evaluated. The
results of experiments performed on the Shelf\&Tote dataset are
presented in Table \ref{table:perf}.

\subsubsection{Generating training data using test data distribution} 
The objective here is to establish an upper bound for the performance
of a detector trained with simulated images. For this purpose, the
object detector is trained with the knowledge of pose distribution in
the test data. This process consists of estimating the density of the
test data with respect to object poses using {\it Kernel Density
  Estimation}, and generating training data according to this
distribution, as follows:

\begin{itemize}
\item Uniformly simulate many scenes using a simulator and record the
  poses for each object in the scene.
\item Weigh each generated scene according to its similarity to test
  data. This is the sum of the number of objects in the scene for
  which the pose matches (rotation difference less than $15^o$ and
  translation difference less than 5cm) at least one pose in their
  corresponding test pose distribution.
\item Normalize the weights to get a probability distribution on the
  sampled poses.
\item Sub-sample the training poses using the normalized probability
  distribution
\end{itemize}
The sampled scenes were used to train a Faster-RCNN detector, which
achieved an accuracy of 69\%.

\subsubsection{Uniformly sampled synthetic data}
This alternative is a popular technique of generating synthetic
data. It uses 3D models of the objects to render their images from
several viewpoints sampled on a spherical surface centered at the
object. The background image corresponded to the {\tt APC} shelf, on
top of which randomly selected objects were pasted at sampled
locations. This process allows to simulate occlusions and mask
subtraction provides the accurate bounding boxes in these cases. The
objects in these images are not guaranteed to have physically
realistic poses. This method of synthetic data generation does not
perform well on the target task, giving a low accuracy of 31\%.

\subsubsection{Generating training data with physics-aware simulation}
The accuracy of 64\% achieved by the proposed physics-aware simulator
is close to the upper bound. By incorporating the knowledge of the
camera pose, resting surface and by using physics simulation, the
detector is essentially over-fitted to the distribution of poses from
which the test data comes, which can be useful for robotic setups.

\vspace{-.1in}
\begin{table}[h]
\centering
\resizebox{\linewidth}{!}{
\begin{tabular}{|a| m{4.6cm}|m{1.8cm} m{0cm}|}
    \hline
    \rowcolor{Gray}
    & {\bf Method} & {\bf Success(IoU$>$0.5)} &\\[1ex]
    \cline{2-4}
    & Team MIT-Princeton \cite{Princeton} (Benchmark) & 75\% &\\[1ex]
    \hline
    \hline
    & Sampled from test data distribution & 69\% &\\[1ex]
    \cline{2-4}
    & Sampled from uniform distribution & 31\% &\\[1ex]
    \cline{2-4}
    & Physics-aware simulation & 64\% &\\[1ex]
    \cline{2-4}
    \multirow{-4}{*}[2.8ex]{\rotatebox[origin=c]{90}{{\bf Simulation}}} & Physics-aware simulation + varying light & {\bf 70}\% &\\[1ex]
    \hline
    \hline
    & Self-learning (2K images) & 75\% &\\[2ex]
    \cline{2-4}
    & Self-learning (6K images) & 81\% &\\[2ex]
    \cline{2-4}
    \multirow{-3}{*}[4.8ex]{\rotatebox[origin=c]{90}{{\bf Self-Learning}}} & Self-learning (10K images) & {\bf 82}\% &\\[2ex]
    \hline
\end{tabular}}
\vspace{-.1in}
\caption{Evaluation on Princeton's Shelf\&Tote dataset \cite{Princeton}}\label{table:perf}
\vspace{-.25in}
\end{table}

The results discussed until now were with respect to a constant
lighting condition. As the dataset grows, then a dip in the
performance is observed. This is expected as the detector overfits
with respect to the synthetic texture, which does not mimic real
lighting condition. This is not desirable, however. To deal with this
issue, the lighting conditions are varied according to the location
and color of the light source. This does resolve the problem to some
extent but the dataset bias still limits performance to an accuracy of
70\%.

On the table-top setup, the detector trained by the physics-based
simulation has a success rate of 78.8\%, as shown in Table
\ref{table:bbox}.

\vspace{-.1in}
\begin{table}[h]
\centering
\resizebox{\linewidth}{!}{
\begin{tabular}{|m{4.6cm}|p{2.2cm} m{0cm}|}
\hline
\rowcolor{Gray}
{\bf Method} & {\bf Success(IoU$>$0.5)} &\\
\hline
Physics aware simulation & 78.8\% &\\
\hline
Self-learning (140 images) & {\bf 90.3\%} &\\
\hline
\end{tabular}}
\vspace{-.1in}
\caption{Detection accuracy on table-top experiments}
\label{table:bbox}
\vspace{-.3in}
\end{table}

\subsection{Evaluating Self-learning}
The self-learning pipeline is executed over the training images in the
{\tt Shelf\&Tote} \cite{Princeton} training dataset to automatically
label them using multi-view pose estimation. The real images are
incrementally added to the simulated dataset to re-train the {\tt
  Faster-RCNN}. This results in a performance boost of 12\%. This
result also outperforms the training process by \cite{Princeton} which
uses approximately 15,000 real images labeled using background
subtraction. The reason that the proposed method outperforms a large
dataset of real training images is mostly because the proposed system
can label objects placed in a clutter.

\begin{figure}[h]
\centering
\includegraphics[width=\linewidth, height=6.5cm,
  keepaspectratio]{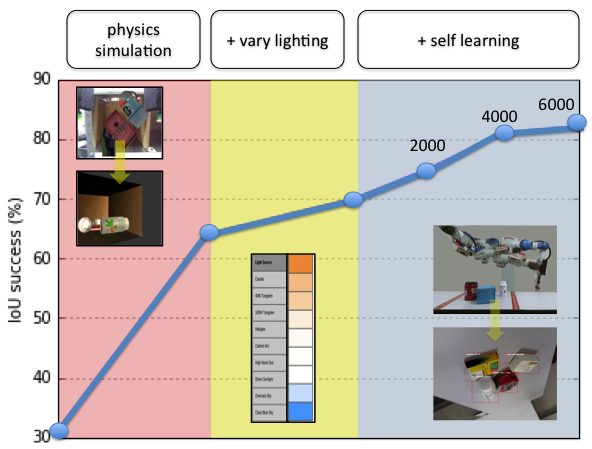}
\vspace{-.25in}
\caption{Plot depicting the performance improvement by adding different components of the system}
\label{fig:improvement}
\vspace{-.2in}
\end{figure}

\begin{figure}[h]
\centering
\includegraphics[width=\linewidth, height=6.5cm,
  keepaspectratio]{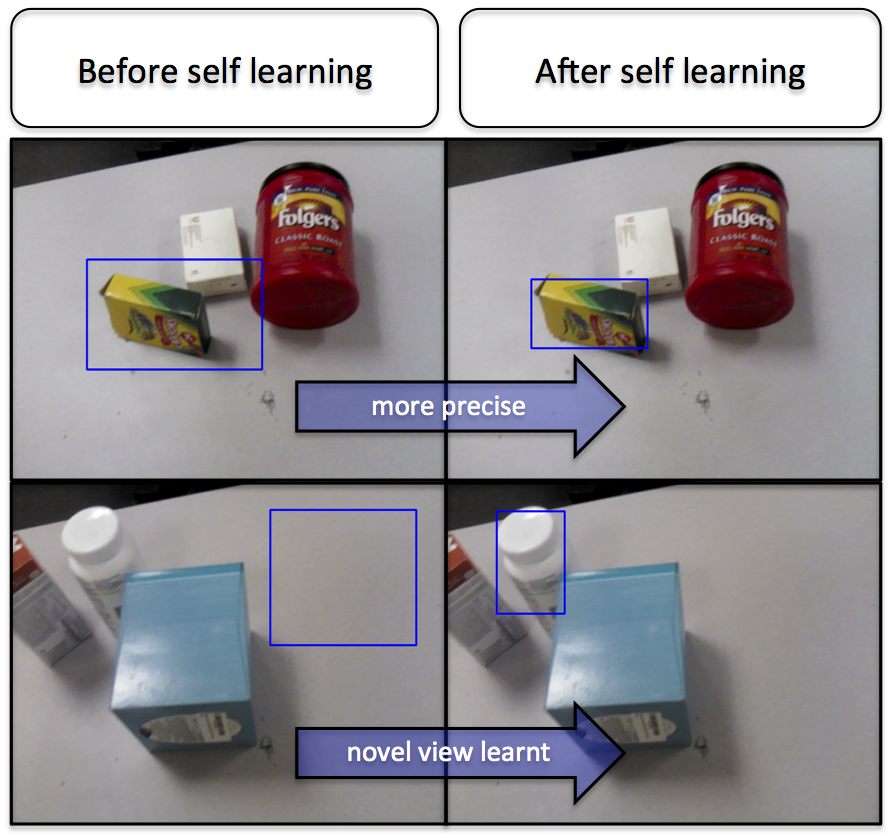}
\vspace{-.15in}
\caption{Results of object detection before and after training with
  the self-learning process. The detector learns to predict more
  precise bounding boxes. It can also detect objects better from novel
  views.}
\label{fig:figurelabel}
\vspace{-.2in}
\end{figure}

On the table-top setup, pose estimation is performed using the trained
detector and model registration. The estimated poses with high
confidence values are then projected to the known camera views to
obtain the 2D bounding box labels on real scenes. This is followed by
reconfiguring the scenes using pick and place manipulation. After
generating 140 scenes with a clutter of 4 objects in each image, the
automatically labeled instances are used to retrain the {\tt
  Faster-RCNN} detector. The performance improvement by adding these
labeled examples is presented in Table \ref{table:bbox}. The overall
performance improvement is depicted in Fig. \ref{fig:improvement},
while an example is shown in Fig.\ref{fig:figurelabel}.

\vspace{-.15in}

\subsection{Evaluating the detector for 6DoF Pose estimation}
Success in pose estimation is evaluated as the percentage of
predictions with an error in translation less than 5cm and mean error
in the rotation less than $15^o$. The results of pose estimation are
compared to the pose system proposed by the APC Team MIT-Princeton
\cite{Princeton} in addition to different model registration
techniques. The results are depicted in Table
\ref{table:poseest}. Given the above specified metric, the proposed
approach outperforms the pose estimation system proposed before
\cite{Princeton} by a margin of 25\%. It is very interesting to note
that the success in pose estimation task is at par with the success
achieved using ground truth bounding boxes.

\vspace{-.15in}

\begin{table*}[t]
    \resizebox{\textwidth}{!}{
    \begin{tabular}{|m{4cm}|m{3cm}|m{3.5cm}|m{3.5cm}|m{2cm} m{0cm}|}
    \hline
    \rowcolor{Gray}
    2D-Segmentation Method & 3D-registration Method & Mean-error Rotation (deg) & Mean-error Translation (m) & Success(\%) &\\[2ex]
    \hline
     Ground-Truth Bounding-Box & PCA + ICP & 7.65 & 0.02 & 84.8 &\\[0.85ex]
    \hline
    \hline
     {\tt FCN} (trained with \cite{Princeton}) & PCA + ICP & 17.3 & 0.06 & 54.6 &\\[0.85ex]
     \hline
     {\tt FCN} (trained with \cite{Princeton}) & Super4PCS + ICP & 16.8 & 0.06 & 54.2 &\\[0.85ex]
     \hline
     {\tt FCN} (trained with \cite{Princeton}) & fast-global-registration & 18.9 & 0.07 & 43.7 &\\[0.85ex]
     \hline
     \hline
     {\tt RCNN} (Proposed training) & PCA + ICP & {\bf 8.50} & 0.03 & {\bf 79.4} &\\[0.85ex]
     \hline
     {\tt RCNN} (Proposed training) & Super4PCS + ICP & 8.89 & {\bf 0.02} & 75.0 &\\[0.85ex]
     \hline
     {\tt RCNN} (Proposed training) & fast-global-registration & 14.4 & 0.03 & 58.9 &\\[0.85ex]
     \hline
    \end{tabular}}
    \vspace{-.1in}
     \caption{Comparing the performance of the proposed system to state-of-the-art techniques for pose estimation.}
     \label{table:poseest}
     \vspace{-.4in}
 \end{table*}

\section{Discussion}
\label{sec:conclusion}
This work provides a system that autonomously generates data to train
{\tt CNNs} for object detection and pose estimation in robotic
setups. Object detection and pose estimation are tasks that are
frequently required before grasping \cite{Azizi:2017aa} or rearranging
objects with a robot \cite{Shuai:2017aa, Krontiris:2015aa}. A key
feature of the proposed system is physical reasoning. In particular,
it employs a physics engine to generate synthetic but physically
realistic images. The images are very similar to real-world scenes in
terms of object pose distributions. This helps to increase the success
ratio of {\tt CNNs} trained with simulated data and reduces the
requirements for manual labeling.

Nevertheless, synthetic data may not be sufficient as they cannot
always generalize to the lighting conditions present in the
real-world. For this purpose and given access to a robotic setup, this
work proposes a lifelong learning approaches for a manipulator to
collect additional labeled data in an autonomous manner. In
particular, the method utilizes successful, high confidence detections
from multiple views to perform pose estimation.  This avoids
over-fitting to simulated conditions. The overall combination of
physical reasoning and self-learning results in a success ratio that
outperforms current state-of-the-art systems in robot vision.

A future objective remains to achieve similar quality of object
detection and pose estimation only with simulated data. This would
minimize the dependency of having access to a robotic setup for
adaptation of the learning procedure to real-world conditions.
Furthermore, it would be interesting, to extend the training process
to facilitate semantic segmentation of scenes, which could lead to an
even more robust pose estimation.

\bibliographystyle{IEEEtran}
\bibliography{physics_perception}

\end{document}